\documentclass[sigconf,nonacm,natbib=true]{acmart}

\AtBeginDocument{%
  \providecommand\BibTeX{{%
    \normalfont B\kern-0.5em{\scshape i\kern-0.25em b}\kern-0.8em\TeX}}}

\setcopyright{acmcopyright}
\copyrightyear{2023}
\acmYear{2023}
\acmDOI{XXXXXXX.XXXXXXX}

\usepackage{color}
\usepackage{multirow}
\usepackage{enumitem}
\usepackage{tcolorbox}
\usepackage{amsmath}
\usepackage{bbold}

\newcommand{\freq}{\mathit{freq}}

\setcopyright{cc}
\copyrightyear{2024}

\begin{document}

\title[{Towards Realistic Synthetic User-Generated Content: A Scaffolding Approach to Generating Online Discussions}]{Towards Realistic Synthetic User-Generated Content:\\ 
A Scaffolding Approach to Generating Online Discussions}

\author{Krisztian Balog}
\authornote{Corresponding author.}
\affiliation{%
  \institution{Google}
  \city{Stavanger}
  \country{Norway}
}
\email{krisztianb@google.com}

\author{John Palowitch}
\affiliation{%
  \institution{Google}
  \city{Mountain View}
  \country{USA}
}
\email{palowitch@google.com}

\author{Barbara Ikica}
\affiliation{%
  \institution{Google}
  \city{Zurich}
  \country{Switzerland}
}
\email{barbaraikica@google.com}

\author{Filip Radlinski}
\affiliation{%
  \institution{Google}
  \city{London}
  \country{UK}
}
\email{filiprad@google.com}

\author{Hamidreza Alvari}
\affiliation{%
  \institution{Google}
  \city{Mountain View}
  \country{USA}
}
\email{hamidrz@google.com}

\author{Mehdi Manshadi}
\affiliation{%
  \institution{Google}
  \city{Mountain View}
  \country{USA}
}
\email{manshadi@google.com}

\renewcommand{\shortauthors}{Balog et al.}

\begin{abstract}
The emergence of synthetic data represents a pivotal shift in modern machine learning, offering a solution to satisfy the need for large volumes of data in domains where real data is scarce, highly private, or difficult to obtain. We investigate the feasibility of creating realistic, large-scale synthetic datasets of user-generated content, noting that such content is increasingly prevalent and a source of frequently sought information. Large language models (LLMs) offer a starting point for generating synthetic social media discussion threads, due to their ability to produce diverse responses that typify online interactions. However, as we demonstrate, straightforward application of LLMs yields limited success in capturing the complex structure of online discussions, and standard prompting mechanisms lack sufficient control. We therefore propose a multi-step generation process, predicated on the idea of creating compact representations of discussion threads, referred to as scaffolds. Our framework is generic yet adaptable to the unique characteristics of specific social media platforms. We demonstrate its feasibility using data from two distinct online discussion platforms. To address the fundamental challenge of ensuring the representativeness and realism of synthetic data, we propose a portfolio of evaluation measures to compare various instantiations of our framework.
\end{abstract}

\ccsdesc[300]{Information systems~Document representation}
\ccsdesc[500]{Computing methodologies~Natural language generation}
\ccsdesc[300]{Computing methodologies~Supervised learning}

\keywords{Synthetic data generation, user-generated content, large language models, evaluation measures}

\maketitle

\section{Introduction}

The emergence of synthetic data represents a pivotal shift in modern machine learning, offering a source of large volumes of data in diverse domains where real data is scarce, highly private, or difficult to obtain. 
Examples include generating images for various computer vision problems~\citep{Nikolenko:2019:arXiv}, question answering corpora for pre-training neural models~\citep{Alberti:2019:ACL}, query logs for evaluating query auto-completion systems~\citep{Krishnan:2020:ECIR}, and dialogue datasets for improving conversational systems~\citep{Kim:2022:EMNLP,Leszczynski:2022:NeurIPS,Leszczynski:2023:arXiv}.
Synthetic data is particularly useful in the realm of user-generated content (UGC), where research and development critically depends on the availability of large-scale datasets that allow for the modeling and analysis of the dynamics of users and content.
There, collecting real data is subject to complex privacy legislation, often expensive to collect, and problematic to distribute or share with other researchers, thereby hindering progress.
In this paper, we investigate the feasibility of the creation of realistic synthetic corpora involving a wide variety of discussions among different users as may be observed across a broad range of user-generated content platforms.

Existing work has explored the generation of both graph and text data for social media data~\citep{Sagduyu:2018:TCSS, qin2017generating, muric2022large, yoon2023graph, park2022social}. However, realistic natural language back-and-forth in conversations common in UGC has only attracted limited attention. In particular, UGC often consists of a sequence of utterances, which progress the topic of a conversation to a greater or lesser degree depending on the topic and participants that the platform attracts. We argue that modeling this structure is key to synthetic collections being usefully representative of actual interactions. We thus focus on generating synthetic collections that accurately represent user dynamics.

Large language models (LLMs) offer a natural starting point for generating synthetic social media discussion threads, given their capabilities to produce diverse responses that typify online interactions.
However, as we demonstrate in this paper, LLMs out of the box have limited success in realistically capturing the complex structure of online discussion threads, and the standard prompting mechanism does not offer enough control over the generation process.
Our solution is a multi-step data generation pipeline, predicated on the idea of creating compact representations of discussion threads, referred to as \emph{scaffolds}.
Our framework is designed to be generic in nature, yet provide the flexibility for individual components to be customized according to the unique characteristics of specific platforms.

While measures exist to compare synthetic and real data in terms of topical distribution, content, and structural properties of discussion threads, none adequately capture the realisticity of a sequence of exchanges. Therefore, we also propose a novel LLM-based \emph{realism} measure. This measure, combined with existing measures, forms our suite of evaluation measures. 
We compare various instantiations of our framework and demonstrate its feasibility using data from two online discussion platforms with markedly different characteristics (Reddit and Wikipedia Talk Pages).

In summary, our main contributions are:
(1) a framework and baseline for generating realistic synthetic UGC conversations;
(2) a novel scaffolded generation approach;
(3) a suite of evaluation measures, including a new LLM-based realism measure;
(4) a comprehensive experimental evaluation demonstrating the feasibility of our approach.

\section{Related Work}

The methods we explore in this paper use LLMs \cite{naveed2023comprehensive} to iteratively construct and refine successive replies in a single online discussion thread, generating both the text in the discussion and the complete discussion structure. Posts form a directed, acyclic graph such that each node has a single parent (either the original post or a reply to the post). Our work is therefore connected to the growing area of graph generation, which can be (roughly) divided into two sub-areas focused on \emph{small graphs} and \emph{large graphs}. 
For small graphs, approaches include neural generative models for molecular structure discovery~\citep{you2018graphrnn,dai2020scalable,liao2019efficient,zahirnia2022micro} and probabilistic generative models for benchmarking graphical neural networks (GNNs)~\citep{palowitch2022graphworld,dwivedi2023benchmarking}.
The main difference between these works and ours is that we do not specify an explicit distribution on graphs: We use LLMs to generate the entire thread, including tokens that indicate the parent comment. 
As in real-world UGC systems, this generates an implicit graph structure, for which we propose evaluations to measure the realism of our generated discussion trees. On the large-graph side, approaches include privacy-aware single-graph generation for benchmarking GNNs on large (privacy-restricted) graph data (such as social networks)~\citep{yoon2023graph,qin2017generating,Darabi:2023:arXiv,Sagduyu:2018:TCSS} and probabilistic and agent-based models for generating large-scale social network data specifically~\citep{Will:2020:SESM,muric2022large,park2022social}. We diverge from this area because we generate individual threads, one-by-one, rather than relying on the asynchronous emergence of threads in the total UGC system. Our focus is therefore on within-thread realism rather than cross-thread interactions.

When generating long or complex text, neural language models have been observed to at times generate repetitive \cite{Suzuki:2017:EACL} or inconsistent \cite{Liu:2022:ICDMW} text. 
This is particularly problematic in our setting, where the back and forth of the online discussion is generally desired to be consistent and on topic. While some recent research has attempted to address this issue through modeling author personas~\cite{Liu:2022:ICDMW}, we focus on the structure of the conversation as a whole.  
Taking inspiration from solutions employed in long-form QA approaches such as 
Chain-of-Thought \cite{Wei:2022:NeurIPS} and Blueprints \cite{Narayan:2023:TACL}, we propose a multi-step process that iteratively refines the content generated to maximize consistency. Closest to blueprints, which generate question-answer pairs that are desirable to answer in long-form generated content \cite{Narayan:2023:TACL}, we generate desirable conversation structure that is progressively specified in increasing detail.

\citet{veselovsky2305generating} propose solutions to increase the faithfulness of LLM-generated synthetic data, where the problem is that the distributions of synthetic and real data often differ~\cite{alaa2022faithful}. Specifically, they study three strategies: (1) grounding, which involves providing real-world examples from a training set in prompts, (2) filtering, which involves leveraging a discriminator model to distinguish real and synthetic data, and (3) taxonomy-based generation, which consists of using a taxonomy in the prompt to increase diversity. 

\citet{li2023synthetic} conduct an evaluation study on the effectiveness of LLM-generated synthetic data w.r.t. \emph{subjectivity} of classification tasks at two levels: (1) \emph{task-level} (i.e., whether this
type of classification task is subjective), and (2) \emph{instance-level} (i.e., whether there are disagreements on the label of this task instance). Their findings suggest that subjectivity at both levels is negatively associated with the performance of the model trained on synthetic data.
\section{Problem Statement}
\label{sec:problem}

We now formalize the problem that we address in detail, showing how it derives from the motivations presented above. 

\subsection{Scoping Synthetic UGC Generation}

As discussed earlier, there are numerous compelling reasons to create synthetic data collections. In many settings key motivators are \emph{data augmentation}, where limited real data is expanded upon to reduce overfitting, and \emph{bias mitigation}, where imbalances exist in real data such as due to poor representation of minority classes. More recently, many authors have observed that synthetic data also improves \emph{user privacy}, avoiding the use of sensitive or personally identifiable data in research settings that do not require it \cite{liu2021machine,el2020practical,boutros2022sface}.
Finally, synthetic data enables \emph{reproducible research} by avoiding time-consuming and costly data collection, which can be particularly challenging for biased and/or sensitive data.
All these issues are important in the context of user-generated content, where data may contain private details, reflect inherent biases within society, and be limited in quantity or diversity.

In this paper, we address the broad challenge of generating realistic synthetic user-generated content. As a first step, we focus on generating online discussion threads given that many platforms encourage or depend on discussions. 
We emphasize demonstrating that this synthetic data can be generated meaningfully from a small amount of real-world training data (i.e., augmentation), while ensuring that diverse online communities are equitably represented (i.e., bias mitigation), leaving privacy protection and downstream utilization for future work.
We also focus on establishing metrics of the alignment between synthetic and real-world data, as these will guide the evaluation and refinement of approaches. Overall, this work lays a strong foundation for future research in this area, enabling further exploration of synthetic data generation and its applications.

\begin{figure*}[t]
    \centering
    \includegraphics[width=0.9\textwidth]{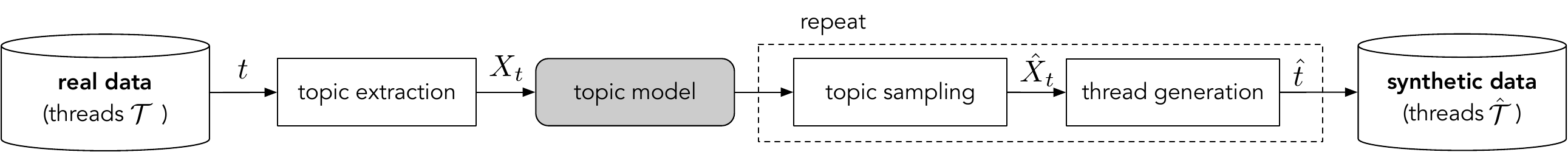}
    \caption{Overview of our synthetic data generation framework. Squared rectangles indicate processes, while rounded rectangles with shaded backgrounds signify models. Approaches for thread generation are detailed in Figure~\ref{fig:thread_generation}.}
    \label{fig:pipeline}
\end{figure*}

\subsection{Problem Definition}
\label{sec:problem_def}

We assume an online platform that facilitates discussion between users. The discussions may be organized across multiple communities $\mathcal{C}$, where each community brings together users who share a common topical interest, cause, or identity.
A discussion thread consists of a set of posts $t=\{p_0,p_1,\dots,p_n\}$.
More specifically, each thread has an \emph{opening post} $p_0$ followed by a sequence of zero or more \emph{replies} $p_i$ ($i \in \{1,\dots, n\}$). When the distinction is not essential, we refer to the opening post and replies collectively as \emph{posts}. 
Replies $p_i$ respond either to the opening post or a previous reply $p_j$ $(0 \leq j < i)$.
Each post $p_i$ is created by a single user. Depending on the platform, users might later edit or remove their posts; however, those actions are currently not modeled.
Thus, each post $p_i$ ($i \in \{0,\dots,n\}$) may be represented as a tuple $p_i=(c_i,u_i,pp_i)$, where $c_i$ is the content of the post, $u_i$ is the author, and $pp_i$ is the parent post (for the opening post $p_0$ the parent is $\emptyset$).

\begin{table}[t]
  \caption{Notation used in the paper.}
  \label{tab:notation}
  \begin{tabular}{ll}
    \toprule
    \textbf{Symbol} & \textbf{Description} \\
    \midrule
    $p_i=(c_i,u_i,pp_i)$ & Post with content $c_i$, author $u_i$, and parent $pp_i$ \\
    $s$ & Scaffold \\
    $\mathcal{S}$ & Set of scaffolds \\
    $t$ & Discussion thread \\
    $\mathcal{T}$ & Set of threads \\
    $x$ & Topic \\
    $X_t$ & Set of topics characterizing thread $t$ \\
    \bottomrule
\end{tabular}
\end{table}

\begin{figure*}[t]
    \centering
    \includegraphics[width=\textwidth]{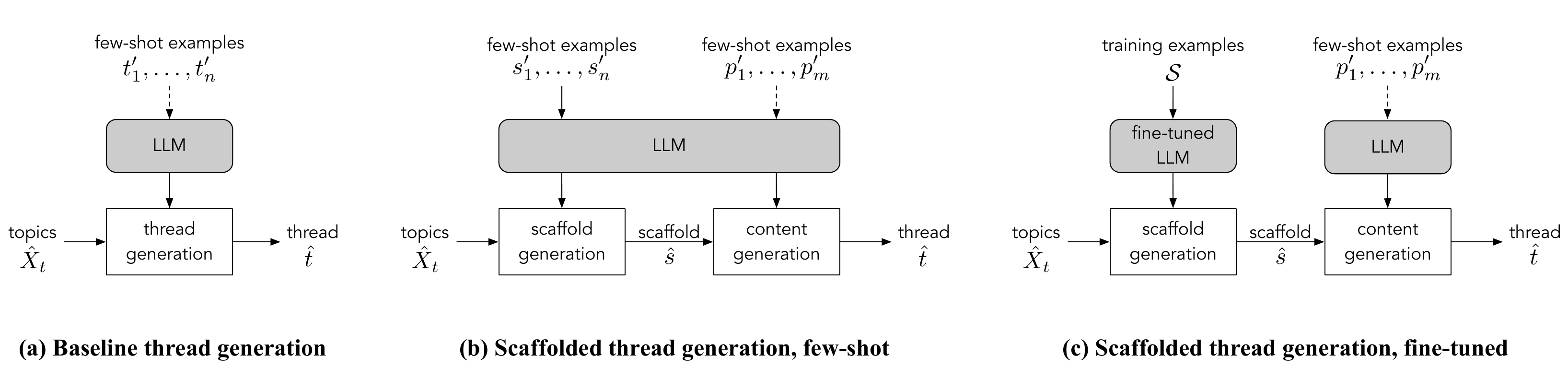}
    \caption{Thread generation approaches. Dashed lines indicate optional input, i.e., the task may be performed either as zero-shot or as few-shot prompting the LLM.}
    \label{fig:thread_generation}
\end{figure*}

Given a sample of real threads $\mathcal{T}$, our objective is to generate a set of synthetic threads $\widehat{\mathcal{T}}$ such that the synthetic threads capture the characteristics of the real threads as measured by some evaluation function. 
In this study, we focus on characteristics related to the structure and content of the threads and discuss specific choices of measures in Section~\ref{sec:evalmeasures}. 
Additional layers of complexity, such as modeling the behavior of users across multiple threads, not just within a thread, and encompassing aspects of temporality, beyond the sequential order of post creation, are left to future work.

\section{Synthetic Data Generation}
\label{sec:framework}

This section presents our framework for generating a collection of synthetic discussion threads based on a sample of real threads.
We propose a multi-step pipeline, shown in Fig.~\ref{fig:pipeline}, that leverages a large language model (LLM) in several key stages. Designed to be model-agnostic, our approach can utilize any pre-trained, instruction-tuned LLM.
First, we extract the main discussion topics from each real thread and model their distribution across the collection. To generate new synthetic threads, we sample a set of topics from this model (Section~\ref{sec:framework:topics}).
We explore two distinct thread generation approaches: a baseline that generates threads directly by few-shot prompting an LLM (Section~\ref{sec:framework:baseline}), and a novel scaffolded approach that first creates a compressed representation of a thread and then populates the content of each post in the thread (Section~\ref{sec:framework:scaffolded}).

Our descriptions below focus on the main ideas behind our methods. For reproducibility, specific prompt examples are provided in Appendix~\ref{app:prompts}.
Throughout, we adhere to the notational consistency of using the hat symbol (~~$\hat{}$~~) for data that is synthetically generated.

\subsection{Topic Extraction and Sampling}
\label{sec:framework:topics}

Our framework starts by extracting one or more topics ($X_t$) from each thread ($t$) in the training data, such that these topics capture the main themes of discussion. 
This task can effectively be performed with the help of LLMs (\S\ref{sec:framework:topics:extraction}).
Next, we build a topic model using all extracted topics to help us generate new synthetic threads, where each discusses a realistic and coherent set of topics. 
We present two approaches for topic sampling: one that samples each topic independently (\S\ref{sec:framework:topics:sampling_ind}) and another that incorporates the relationship between topics in a pairwise manner (\S\ref{sec:framework:topics:sampling_cond}).
These approaches resemble unigram and bigram language models, with topics treated akin to terms.

\subsubsection{Topic Extraction}
\label{sec:framework:topics:extraction}

To extract topics for a thread, we prompt a pre-trained LLM with few-shots demonstrators, i.e., manually curated examples of threads and corresponding topics (see Appendix~\ref{app:prompts:topics} for the actual prompt used). 
The prompt includes the text from all posts from the thread (starting with the initial post), in order of posting, up to the LLM's context limit.

\subsubsection{Independent Topic Sampling}
\label{sec:framework:topics:sampling_ind}

Let $\theta_l$ be the distribution of \emph{topic lengths}, where length refers to the number of topics that are extracted for a given thread. Specifically, $P(l|\theta_l)$ is proportional to the number of threads in the training data for which $l$ number of topics are extracted:
$P(l|\theta_l) = \big|\{t \in \mathcal{T} : |X_t| = l \}\big| / |\mathcal{T}|$.

Let $\theta_x$ be a \emph{topic distribution}, such that $P(x|\theta_x)$ is proportional to the number of occurrences of $x$ across all topics identified in all threads:
$P(x|\theta_x) = \big|\{t \in \mathcal{T}: x \in X_t \}\big| / \sum_{t' \in \mathcal{T}}|X_{t'}|$, 
where the numerator is the total number of training threads labeled with $x$.

With these probability distributions estimated from training data, a set of topics $\hat{X}_t$ (to characterize the synthetic thread) is sampled using the following algorithm:
\begin{enumerate}
    \item Sample the number of topics: $m \sim \theta_l$
    \item $\hat{X}_t = \emptyset$
    \item While $|\hat{X}_t| < m$:
    \begin{enumerate}[label=(\roman*)]    
        \item Sample a new topic $x \sim \theta_x$
        \item $\hat{X}_t = \hat{X}_t \cup \{x\}$
    \end{enumerate}
\end{enumerate}

\subsubsection{Conditional Topic Sampling}
\label{sec:framework:topics:sampling_cond}

Assuming independence between topics in one discussion is clearly an oversimplification and might result in a set of topics that are extremely unlikely to be discussed together. 
Therefore, we present an improved topic sampling approach that also considers how frequently any given pair of topics is discussed together.

Let $\theta_{x'|x}$ be a \emph{conditional topic distribution}, where $P(x'|x, \theta_{x'|x})$ expresses the likelihood of topic $x'$ being extracted for a thread that also has $x$ as a topic ($x' \neq x$): 
\begin{equation}
    P(x'|x,\theta_{x'|x}) = \frac{ \freq(x,x') + 1 } { \sum_{x''}\freq(x,x'') + M - 1 } ~,    
\end{equation}
where $\freq(x,x')$ is the total number of training threads that are labeled with both $x$ and $x'$, and $M$ is the total number of unique topics.
Notice that we apply Laplacian smoothing to ensure that topics that do not co-occur with any other topics in the training data still have have a non-zero probability of being selected.

The process of sampling a set of topics $\hat{X}_t = \{x_1,\dots,x_m\}$ to characterize the synthetic thread (to be generated) is as follows:
\begin{enumerate}
    \item Sample the number of topics: $m \sim \theta_l$
    \item Sample the first topic: $x \sim \theta_x$; $\hat{X}_t = \{x\}$
    \item While $|\hat{X}_t| < m$:
    \begin{enumerate}[label=(\roman*)]
        \item Randomly pick from the topics sampled so far: $x \sim \hat{X}_t$
        \item Conditionally sample the next topic: $x' \sim \theta_{x'|x}$
        \item $\hat{X}_t = \hat{X}_t \cup \{x'\}$
    \end{enumerate}
\end{enumerate}

\subsection{Baseline Thread Generation}
\label{sec:framework:baseline}

Given a set of input topics $\hat{X}_t$, we prompt the LLM to generate a discussion thread $\hat{t}$, requesting that it adheres to a particular format that encodes the internal structure of the thread.
Specifically, a thread discussing a set of topics is represented as follows:
\begin{small}
\begin{verbatim}
  topics: <topic_1>[, <topic_i>]*
  title: <initial post title>
  [<post_ID> # <user_ID> # <parent_ID> # <post content>]+
\end{verbatim}
\end{small}
In this format, the parent ID field references the ID of the post to which the current post is directly replying; for the opening post, this field is set to \texttt{NA}.

To generate a new thread, we provide the LLM with zero or more examples of existing threads (corresponding to zero-shot and few-shot settings), encoded in the above format along with the set of topics to be discussed (see Appendix~\ref{app:prompts:baseline} for the prompt used). To ensure diversity and realism, we randomly sample new examples from real threads for each synthetic thread, as opposed to using manually curated demonstrations.  We then instruct the LLM to generate a thread in the specified format, starting from the second line (i.e., the initial post).
It is important to note that there's no guarantee the LLM's output will perfectly adhere to this format or maintain a valid internal structure (e.g., with correct parent IDs).

\subsection{Scaffolded Thread Generation}
\label{sec:framework:scaffolded}

A key idea in this paper is to provide more control over the thread generation process by performing it in two steps: (1) generating a compact representation, called \emph{scaffold}, that encodes the structure of the discussion along with the summary of each post, and (2) generating the actual content of each post, based on its summary.
The scaffold is encoded using the same multi-line fielded structure as described above, the main difference being that the last field for each post is a \emph{summary} of the post instead of its actual content.
For each real thread, a scaffold is created with a summary of each post (\S\ref{sec:framework:scaffolded:examples}); Figure~\ref{fig:scaffold} shows an example.
These scaffolds are then used for the generation of synthetic scaffolds, either as few-shot examples in a prompt (\S\ref{sec:framework:scaffolded:fewshot}) or as training instances for further fine-tuning an LLM for this particular task  (\S\ref{sec:framework:scaffolded:finetuned}).
In both cases, the summaries of each post in the scaffold are used to prompt the LLM to generate the corresponding post content (\S\ref{sec:framework:scaffolded:content}); see Figs.~\ref{fig:thread_generation} (b) and (c).

\subsubsection{Creating Example Scaffolds}
\label{sec:framework:scaffolded:examples}

Scaffolds that can be used as training examples, either in few-shot prompts or for supervised fine-tuning, are generated from real threads by instructing the LLM to summarize the content of each post; see Appendix~\ref{app:prompts:summaries}.

\begin{figure}[t]
\begin{footnotesize}
\begin{verbatim}
topics: Nonprofit Management, Paint Color, Curtains, Bar Decor, 
    Wallpaper, Countertop
title: Help with decorating!
post # user-1 # NA # The user is looking for suggestions on how
    to decorate the small bar area in their non-profit office.
comment-1 # user-2 # post # The user suggests that the poster may
    want to remove the popcorn machine.
comment-2 # user-1 # comment-1 # The user is not willing to remove 
    the popcorn machine as it was a gift from a well-wisher.
comment-3 # user-3 # post # The user finds the curtains to be a 
    great find and suggests that they be cleaned and hung to look 
    like new. They also suggest painting the countertop.
comment-4 # user-1 # comment-3 # The user expresses gratitude to 
    the commenter for tips on enhancing their bar area.
\end{verbatim}
\end{footnotesize}
    \caption{Example scaffold.}
    \label{fig:scaffold}
\end{figure}

\subsubsection{Few-shot Scaffold Generation}
\label{sec:framework:scaffolded:fewshot}

Our first approach builds upon the baseline thread generation strategy. As before, we prompt the LLM with a set of input topics and randomly sampled few-shot examples from the real dataset (see Appendix~\ref{app:prompts:scaffolds}). However, the examples here are thread scaffolds, not full threads. 
This approach provides more control over the generation of post content, which is delegated to a subsequent step.
Similar to the baseline, this approach may yield invalid scaffolds.

\subsubsection{Fine-tuned Scaffold Generation}
\label{sec:framework:scaffolded:finetuned}

In a second approach, we investigate fine-tuning an LLM directly on thread scaffolds generated from real data. This fine-tuning strategy focuses on improving the degree to which the model internalizes the valid structure of thread scaffolds, increasing the likelihood well-formed outputs. To achieve this, we need to incorporate scaffold validity in the training such that the model is penalized for structurally incorrect outputs.

In supervised fine-tuning for text generation tasks, one way to guide the model's learning is to compute an evaluation metric on the validation set at regular training intervals. A commonly used metric for this purpose is ROUGE, which measures the similarity between generated text and reference text. Here, the scaffolds corresponding to real threads serve as reference.
We use the following augmented metric which also incorporates structural validity:
\begin{equation}
    VM(\mathcal{\hat{S}}_v, \mathcal{S}_v) =  \frac{1}{|\mathcal{\hat{S}}|} \sum_{\hat{s} \in \mathcal{\hat{S}}_v}  \big( \alpha R(s, \hat{s}) + (1 - \alpha) \mathbb{1}(\hat{s}) \big) ,
\end{equation}
where $\mathcal{\hat{S}}_v$ and $\mathcal{S}_v$ are the generated and real scaffolds in the validation set, respectively, $R(s, \hat{s})$ is a ROUGE variant, $\mathbb{1}(\hat{s})$ is a binary measure of scaffold validity, and $\alpha$ is a weighting factor.

\subsubsection{Content Generation}
\label{sec:framework:scaffolded:content}

Once a scaffold for a thread has been created, the next step is to replace the brief summaries of each individual post with text that is both realistic and aligns with the summary. Importantly, realistic text does not always imply length or complexity. Depending on the context, realistic text could range from a single emoji to more elaborate expressions.

The content of the first post of a thread ($c_0$) is generated by prompting the LLM with just the topics and required first post summary.  For all later posts, their contents $c_i$ are generated in scaffold sequence, one by one, while enabling the LLM to recognize the back-and-forth conversation that preceded the post being written; see Appendix~\ref{app:prompts:content}.

Specifically, the content of some post $p_i$ is very likely to depend on the post to which it is a direct response, $pp_i$, so this must be provided as context to the LLM. However, $p_i$ may also, in principle, depend on any earlier post in the discussion (which the author may have read). The LLM generating $p_i$ could thus in principle require all earlier posts to be provided as context. Long contexts are known to reduce LLM performance,\footnote{Long context LLMs are an active area of research with significant progress, meaning that all earlier posts \emph{could} be provided as context, even for long threads. However, there are a number of downsides, including computational cost.} thus we assume the necessary context is only the direct chain of parent posts from $p_i$ up to the first post of the thread. This sequence of posts and direct replies leading to $p_i$ is thus provided as context to the LLM along with the thread topics and post summary from the scaffold.
\section{Evaluation Measures}
\label{sec:evalmeasures}

Our objective is to generate a set of synthetic threads ($\widehat{\mathcal{T}}$) that are valid (Section~\ref{sec:evalmeasures:validity}) and characteristically similar to real threads ($\mathcal{T}$).
We present evaluation measures to capture this similarity in terms of topics covered (Section~\ref{sec:evalmeasures:topic}), thread structure (Section~\ref{sec:evalmeasures:structure}), and content (Section~\ref{sec:evalmeasures:content}). Additionally, we develop a novel \emph{realism} measure to quantify the coherence of the dialogue in a thread (Section~\ref{sec:evalmeasures:real}).

\subsection{Validity}
\label{sec:evalmeasures:validity}

A valid structure means that a thread has an opening post and each of the replies respond to a previous post (cf. Section~\ref{sec:problem_def}). Validity might be violated if the LLM-generated output is synthetically incorrect (e.g., a field like user or content is missing) or if a post replies to a non-existing (future) post.
We define \emph{success rate} as the portion of threads that have a valid structure. 

\subsection{Topic Measures}
\label{sec:evalmeasures:topic}

To compare a set of synthetic threads to a set of real threads at the topic level, we first extract topics from each thread in each set. We then produce a topic-probability vector $\mathbf{v}$ for each set, where $\mathbf{v}[x]$ represents the proportion of topic $x$ in the given set. Our goal is to compare $\mathbf{v}$ and $\mathbf{\hat{v}}$, corresponding to the real and synthetic thread sets, respectively. To do this, we use two metrics:
\begin{enumerate}
    \item \textbf{Jensen-Shannon divergence}~\citep{menendez1997jensen}. We compute the J-S divergence of the probability vectors.
    \item \textbf{Weighted Jaccard similarity}. We compute the weighted Jaccard similarity, defined as
    \begin{equation}
        \text{W-J}(\mathbf{\hat{v}}, \mathbf{v}) = \sum_i\dfrac{\min(\mathbf{\hat{v}}[x], \mathbf{v}[x])}{\max(\mathbf{\hat{v}}[x], \mathbf{v}[x])}.
    \end{equation}
\end{enumerate}
Mind the notion of topics is the same as in Sec.~\ref{sec:framework:topics}, but they are obtained differently here in order for evaluation to be independent of the data generation process.
Specifically, we use Google's Cloud Natural Language Inference toolkit\footnote{\url{https://cloud.google.com/natural-language/docs/classifying-text}} as the topic classification model. 
Given the text of a thread, it returns a list of topics from a pre-specified, finite list, with entries like \texttt{/News/Politics} and \texttt{/Health/Public Health}, along with confidence scores for each. We retain topics with confidence scores $\geq 0.1$.

\subsection{Structural Measures}
\label{sec:evalmeasures:structure}

To evaluate scaffolds, we take a closer look at the graphs underlying the discussion threads. Recall that a discussion thread can be represented as a rooted directed acyclic graph, with the root corresponding to the opening post $p_0$, its descendant nodes corresponding to the rest of the posts $p_i$ ($1 \leq i \leq n$), and edges pointing from the posts to their replies. Note that the structure of the graph is not modeled explicitly but is generated implicitly, as the LLM constructs entire discussion threads.

To assess how realistic the graphs underlying synthetically generated discussion threads are, we focus on metrics capturing graph structure as well as metrics encapsulating user-posting behavior, where every user is assumed to participate in a single discussion thread only.
Specifically, we consider the following metrics evaluating graph structure:
\begin{enumerate}[label=(\alph*)]
    \item number of posts,
    \item number of unique users participating in the thread,
    \item maximum depth (i.e., length of the longest reply chain),
    \item maximum breadth (i.e., the max. number of posts at one depth),
    \item \textbf{Wiener index} \citep{Wiener:1947:JACS} (i.e., the sum of the lengths of the shortest paths between all (unordered) pairs of posts):
    $$\mathcal{W} = \frac{1}{2} \sum_{i = 0}^n \sum_{j = 0}^n d(p_i, p_j),$$
    \item \textbf{structural virality} \citep{Goel:2016:ManageSci} (i.e., the average shortest path length between all (unordered) pairs of posts):
    $$\mathcal{V}_S = \frac{1}{\binom{n}{2}} \mathcal{W},$$
    \item \textbf{cascade virality} \citep{Zhang:2020:EPL} (i.e., the sum of average shortest path lengths between all posts $p_i$ and their (in)direct replies $R_i$):
    $$\mathcal{V}_C = \sum_{i = 0}^n \sum_{j \in R_i} \frac{d(p_i, p_j)}{|R_i|}.$$
\end{enumerate}
     
We also consider metrics evaluating within-thread user-posting behavior:
(a) number of posts posted by the user,
(b) mean depth at which the user posted,
(c) mean number of direct replies to the user, and
(d) mean number of (in)direct replies to the user.\footnote{As these are not key to our conclusions, the results are reported in Appendix~\ref{app:user_measures}.}

\begin{table*}[t]
    \caption{Thread generation results on the Reddit dataset in terms of success rate and structural measures. Measures are macro-averages over subreddits, with standard deviations shown in parentheses. The first row reports these statistics on the test set for reference comparison.}
    \label{tbl:res:reddit:struct}
    \resizebox{\textwidth}{!}{%
    \begin{tabular}{l|l|c|ccccccc}
    \toprule
    \textbf{Topic} & \textbf{Thread} & \textbf{Success} &  \multicolumn{7}{c}{\textbf{Structural measures}} \\
    \textbf{sampl.} & \textbf{generation} & \textbf{rate} & \textbf{\#posts} & \textbf{\#uniq. users} & \textbf{Max depth} & \textbf{Max breadth} & \textbf{$\mathcal{W}$} & \textbf{$\mathcal{V}_S$} & \textbf{$\mathcal{V}_C$} \\
    \midrule
        n/a & TEST SET & n/a & 14.95 ($\pm 17.82$) & 8.81 ($\pm 8.21$) & 3.74 ($\pm 1.63$) & 6.31 ($\pm 7.03$) & 19,815.82 ($\pm 146,788.06$) & 1.97 ($\pm 0.87$) & 12.55 ($\pm 21.92$) \\
    \midrule
        \multirow{3}{*}{IND}
            & BASELINE & 0.66 ($\pm 0.14$) & 8.99 ($\pm 2.75$) & 8.08 ($\pm 2.89$) & 3.48 ($\pm 0.59$) & 5.51 ($\pm 2.51$) & 137.37 ($\pm 85.18$) & 2.06 ($\pm 0.22$) & 5.24 ($\pm 2.43$) \\
            & SCAFFOLD/FewS & 0.95 ($\pm 0.04$) & 8.69 ($\pm 3.16$) & 7.97 ($\pm 3.15$) & 3.96 ($\pm 0.53$) & 4.18 ($\pm 2.15$) & 143.87 ($\pm 107.25$) & 2.23 ($\pm 0.33$) & 6.94 ($\pm 2.57$) \\
            & SCAFFOLD/FineT & 0.96 ($\pm 0.07$) & 8.79 ($\pm 4.98$) & 6.38 ($\pm 3.94$) & 3.81 ($\pm 1.31$) & 3.98 ($\pm 2.46$) & 270.55 ($\pm 273.99$) & 2.11 ($\pm 0.70$) & 7.36 ($\pm 6.18$) \\
    \midrule
        \multirow{3}{*}{COND}
            & BASELINE & 0.66 ($\pm 0.13$) & 9.14 ($\pm 2.75$) & 8.21 ($\pm 2.88$) & 3.49 ($\pm 0.58$) & 5.62 ($\pm 2.55$) & 141.73 ($\pm 86.74$) & 2.07 ($\pm 0.21$) & 5.28 ($\pm 2.42$) \\
            & SCAFFOLD/FewS & 0.95 ($\pm 0.04$) & 8.80 ($\pm 3.22$) & 8.07 ($\pm 3.19$) & 3.97 ($\pm 0.52$) & 4.23 ($\pm 2.18$) & 149.51 ($\pm 112.23$) & 2.24 ($\pm 0.33$) & 7.06 ($\pm 2.65$) \\
            & SCAFFOLD/FineT & 0.95 ($\pm 0.07$) & 9.09 ($\pm 5.07$) & 6.54 ($\pm 3.97$) & 3.89 ($\pm 1.38$) & 4.08 ($\pm 2.46$) & 287.57 ($\pm 284.46$) & 2.16 ($\pm 0.72$) & 7.75 ($\pm 6.83$) \\
    \bottomrule
    \end{tabular}
    }
\end{table*}

\begin{table}[t]
    \caption{Thread generation results on the Reddit dataset in terms of topic, content, and realism. Measures are macro-averages over subreddits, with standard deviations shown in parentheses. For reference, the first row compares the training data against the test set.}
    \label{tbl:res:reddit:content}
    \resizebox{\columnwidth}{!}{%
    \begin{tabular}{l|l|cc|c|c}
    \toprule
    \textbf{Topic} & \textbf{Thread} & \multicolumn{2}{c|}{\textbf{Topic measures}} & \textbf{Content} & \multirow{2}{*}{\textbf{Realism}} \\
    \textbf{sampl.} & \textbf{generation} & \textbf{J-S} & \textbf{W-J} & \textbf{MAUVE} \\
    \midrule
        n/a & TRAIN SET & 0.76 ($\pm 0.09$) & 0.72 ($\pm 0.13$) & 0.51 ($\pm 0.18$) & 0.97 ($\pm 0.04$)\\
    \midrule
        \multirow{5}{*}{IND}
            & BASELINE & 0.68 ($\pm 0.08$) & 0.55 ($\pm 0.12$) & 0.29 ($\pm 0.12$) & 0.96 ($\pm 0.03$)\\ 
            & SCAFFOLD/FewS + CONT/ZeroS & 0.65 ($\pm 0.09$) & 0.51 ($\pm 0.13$) & 0.27 ($\pm 0.12$) & 0.97 ($\pm 0.02$)\\
            & SCAFFOLD/FewS + CONT/FewS & 0.67 ($\pm 0.09$) & 0.54 ($\pm 0.13$) & 0.28 ($\pm 0.13$) & 0.97 ($\pm 0.03$)\\
            & SCAFFOLD/FineT + CONT/ZeroS & 0.69 ($\pm 0.08$) & 0.57 ($\pm 0.11$) & 0.34 ($\pm 0.16$) & 0.86 ($\pm 0.09$)\\
            & SCAFFOLD/FineT + CONT/FewS & 0.71 ($\pm 0.08$) & 0.60 ($\pm 0.11$) & 0.37 ($\pm 0.16$) & 0.87 ($\pm 0.07$)\\
    \midrule
        \multirow{5}{*}{COND}
            & BASELINE & 0.67 ($\pm 0.12$) & 0.55 ($\pm 0.12$) & 0.28 ($\pm 0.12$) & 0.96 ($\pm 0.03$) \\ 
            & SCAFFOLD/FewS + CONT/ZeroS & 0.65 ($\pm 0.09$) & 0.51 ($\pm 0.13$) & 0.27 ($\pm 0.12$) & 0.97 ($\pm 0.03$)\\
            & SCAFFOLD/FewS + CONT/FewS & 0.67 ($\pm 0.09$) & 0.54 ($\pm 0.13$) & 0.29 ($\pm 0.13$) & 0.96 ($\pm 0.03$)\\
            & SCAFFOLD/FineT + CONT/ZeroS & 0.69 ($\pm 0.08$) & 0.57 ($\pm 0.12$) & 0.35 ($\pm 0.16$) & 0.85 ($\pm 0.09$)\\
            & SCAFFOLD/FineT + CONT/FewS & 0.71 ($\pm 0.07$) & 0.60 ($\pm 0.11$) & 0.38 ($\pm 0.16$) & 0.87 ($\pm 0.07$)\\
    \bottomrule
    \end{tabular}
    }
\end{table}

\subsection{Content Measure}
\label{sec:evalmeasures:content}

To compare how close synthetic threads are to real threads in terms of their content, we leverage \textbf{\textsc{Mauve}}~\citep{pillutla2021mauve}, a distributional measure that estimates the gap between a machine-generated distribution $Q$ and a distribution $P$ of human text using Kullback-Leibler (KL) divergence and Type I/II errors defined as follows: 
(Type I) $Q$ puts high mass on text which is unlikely under $P$, and
(Type II) $Q$ cannot generate text which is plausible under $P$.
As shown in~\citep{pillutla2021mauve}, Type I and II errors are essentially $KL(Q|P)$ and $KL(P|Q)$. \textbf{\textsc{Mauve}}\textit{(P, Q)} is then defined as the area under the divergence curve $C(P, Q)$ and lies in $(0, 1]$ (where larger values mean $Q$ is closer to $P$): 
\begin{equation}
    C(P, Q)=\{(\exp(-cKL(Q\mid R_\lambda)),\exp(-cKL(P\mid R_\lambda))\} ~,
\end{equation}
where  $c > 0$ is a hyperparameter for scaling, $R_\lambda=\lambda P + (1-\lambda)Q$ is a mixture distribution and $\lambda \in \Lambda=\{1/n, 2/n, ..., (n-1)/n\}$ is a mixture weight varied between 0 and 1 to compute the uniformally discretized divergence curve over $n$ points. We use the default values of $c=5$ and $n=32$~\citep{pillutla2021mauve}.

We obtain $P$ and $Q$ from $\mathcal{T}$ and $\mathcal{\widehat{T}}$ respectively, by turning their corresponding threads into an embedding space. Specifically, each thread is converted to 768-dimensional embedding vectors by (1) splitting the entire content into sentences of 12 tokens, (2) encoding each using a pre-trained T5 base model~\citep{raffel2020exploring} with a 12-layer transformer architecture, and (3) concatenating the embeddings from all sentences. 

\subsection{Realism Measure}
\label{sec:evalmeasures:real}

A significant limitation of the metrics introduced thus far is their invariance to post content permutation. In other words, if the posts that make up a discussion thread were shuffled, the current set of metrics would not be affected.
Motivated by this, we develop a ``realism'' evaluation setting that attempts to quantify the coherence of the dialogue in a thread. We ground our evaluation at the level of a \emph{discussion path} between two posts $p_i\rightarrow p_j$. 
For any given set of threads $\mathcal{T}$, we sample $N=100$ threads and $M=5$ paths from each thread, uniformly-at-random over paths of up-to length 4. We then convert each thread into a structured discussion string, and prompt an LLM with instructions to evaluate the 
``coherence'' of the discussion. We provide ten few-shot examples (five coherent and five incoherent discussions). 
The realism score for $\mathcal{T}$ is then
\begin{equation*}
\mathrm{Real}(\mathcal{T}) = \dfrac{1}{NM}\sum_{t_i:i\in[N]}\sum_{j\in[M]}\mathbb{1}(\text{LLM declares $p_i 
\rightarrow p_j$ coherent}).
\end{equation*}
We perform a validation experiment for this measure, which is reported, along with the LLM prompt we use, in Appendix~\ref{app:realism}.

\section{Experimental Setup}
\label{sec:expsetup}

This section introduces our research questions, presents the datasets we use in our experiments, provides details on how we generate data using LLMs, and describes our evaluation methodology.

\subsection{Research Questions}

With our experiments, we seek to answer the following questions:

\begin{itemize}
    \item[RQ1] How effective are our approaches at generating synthetic discussion threads that are similar to real ones in terms of topic distribution, thread structure, content, and realism?
    \item[RQ2] Which of the topic sampling approaches (independent or conditional) is more effective?
    \item[RQ3] How effective are standard LLMs at generating a valid thread structure? Does fine-tuning improve performance?
    \item[RQ4] How well do our approaches generalize (1) across different communities within a given setting and (2) across different types of online communities?
\end{itemize}

\subsection{Data}

We conduct experiments on two online discussion platforms with markedly different characteristics: Reddit and Wikipedia Talk Pages.

\subsubsection{Reddit}
We use a small subset of the Pushshift dataset \citep{Baumgartner:2020:ICWSM} hosted on Zenodo~\citep{Baumgartner:2020:Zenodo} under a Creative Commons Attribution 4.0 license. It consists of all Reddit submissions (opening posts) and comments (reply posts) during April 2019. 
The different communities on Reddit, known as subreddits, exhibit a rich variety of unique (often niche) interests, language styles, community norms and expectations. This diversity makes Reddit a valuable testbed to evaluate the adaptability of our generation process to create synthetic discussions that accurately reflect the characteristics of different communities.
To arrive at a diverse and representative set of communities, we perform a form of stratified sampling to select 250 unique subreddits that contain a total of 256K threads. Our sampling and data preprocessing process is detailed in Appendix~\ref{app:dataset:reddit}.

\subsubsection{Wikipedia Talk Pages}

We also use the publicly available WikiConv dataset~\citep{Hua:2018:EMNLP},\footnote{\url{https://github.com/conversationai/wikidetox/tree/main/wikiconv}} which contains all conversations (between contributors to Wikipedia) from the Wikipedia Talk Pages system. There, editors typically participate in focused discussion about the content of (or edits on) a single Wikipedia article. This gives us another unique testbed for our approach: not only does our model need to synthesize new dialogue on the particular topic, it must also synthesize the goal-oriented rhetoric that Wikipedia editors use to collaboratively improve articles.
Due to the vast majority of threads being ``trivial'' (initial post followed by one reply), we use stratified sampling by thread size, resulting in a balanced dataset of 8.1K threads.
The details of data preprocessing and sampling can be found in Appendix~\ref{app:dataset:wikitalk}.

\begin{table*}[t]
    \caption{Thread generation results on the Wikipedia Talk Pages dataset in terms of success rate and structural measures. The first row reports these statistics on the test set for reference comparison.}
    \label{tbl:res:wikitalk:struct}
    \resizebox{0.7\textwidth}{!}{%
    \begin{tabular}{l|l|c|ccccccc}
    \toprule
    \textbf{Topic} & \textbf{Thread} & \textbf{Success} &  \multicolumn{7}{c}{\textbf{Structural measures}} \\
    \textbf{sampl.} & \textbf{generation} & \textbf{rate} & \textbf{\#posts} & \textbf{\#uniq. users} & \textbf{Max depth} & \textbf{Max breadth} & \textbf{$\mathcal{W}$} & \textbf{$\mathcal{V}_S$} & \textbf{$\mathcal{V}_C$} \\
    \midrule
        n/a & TEST SET & n/a & 12.68 & 3.88 & 5.67 & 4.67 & 727.01 & 2.99 & 15.93 \\
    \midrule
        \multirow{2}{*}{IND}
            & BASELINE & 0.54 & 8.06 & 4.61 & 4.26 & 4.16 & 110.98 & 2.23 & 7.41 \\
            & SCAFFOLD/FewS & 0.75 & 12.21 & 7.97 & 5.87 & 5.07 & 322.57 & 3.00 & 15.82 \\
    \midrule
        \multirow{2}{*}{COND}
            & BASELINE & 0.57 & 7.43 & 4.23 & 4.47 & 3.42 & 95.86 & 2.19 & 7.65 \\
            & SCAFFOLD/FewS & 0.75 & 12.08 & 8.00 & 5.64 & 5.18 & 321.16 & 2.94 & 15.50 \\
    \bottomrule
    \end{tabular}
    }
\end{table*}

\begin{table}[t]
    \caption{Thread generation results on the Wikipedia Talk Pages dataset in terms of topic, content, and realism. For reference, the first row compares the training data against the test set.}
    \label{tbl:res:wikitalk:content}
    \resizebox{\columnwidth}{!}{%
    \begin{tabular}{l|l|cc|c|c}
    \toprule
    \textbf{Topic} & \textbf{Thread} & \multicolumn{2}{c|}{\textbf{Topic}} & \textbf{Content} & \textbf{Realism} \\
    \textbf{sampl.} & \textbf{generation} & \textbf{J-S} & \textbf{W-J} & \textbf{MAUVE} &  \\
    \midrule
        n/a & TRAIN SET & 0.86 & 0.83 & 0.99 & 0.98 \\
    \midrule
        \multirow{3}{*}{IND}
            & BASELINE & 0.70 & 0.58 & 0.64 & 0.93 \\
            & SCAFFOLD/FewS + CONT/ZeroS & 0.72 & 0.60 & 0.72 & 0.95 \\
            & SCAFFOLD/FewS + CONT/FewS & 0.73 & 0.64 & 0.70 & 0.93 \\
    \midrule
        \multirow{3}{*}{COND}
            & BASELINE & 0.69 & 0.55 & 0.66 & 0.94 \\
            & SCAFFOLD/FewS + CONT/ZeroS & 0.72 & 0.59 & 0.73 & 0.90 \\
            & SCAFFOLD/FewS + CONT/FewS & 0.72 & 0.61 & 0.73 & 0.89 \\
    \bottomrule
    \end{tabular}
    }
\end{table}

\subsection{Large Language Models}
\label{sec:expsetup:llms}

Our approaches require an instruction fine-tuned LLM. In our experiment, we use PaLM 2~\citep{Palm2:2023:arXiv} in model size S.
The temperature is set to 0.7.
For prompting, we employ the following general structure, where the parts in square brackets are instructions and the elements in curly brackets are input variables:
\begin{footnotesize}
\begin{verbatim}
  [INTRO]
  <for each example_i>
    [EXAMPLE_PREFIX]
    [EXAMPLE_INPUT_PREFIX] {example_input_i}
    [EXAMPLE_OUTPUT_PREFIX] {example_output_i}
  [INPUT_TASK_PREFIX]
  [INPUT_PREFIX] {input_text}
  [OUTPUT_PREFIX]    
\end{verbatim}
\end{footnotesize}
This general structure works both with and without examples, corresponding to the few-shot and zero-shot settings, respectively. The specific prompts used, which are all instantiations of this patterns, are presented in Appendix~\ref{app:prompts} for reproducibility.

Our default setting is to use two examples, which are sampled randomly from the training data each time the model is prompted generate an output.
Exemptions to this are topic extraction (cf. \S\ref{sec:framework:topics:extraction}) and post summarization as part of example scaffold creation (cf. \S\ref{sec:framework:scaffolded:examples}); there, examples cannot be directly sampled from the dataset and have to be manually created.
The input context length is limited to 4K tokens. We ensure that truncation is done along meaningful units for the given task, e.g., posts or sentences.

We fine-tune a single scaffold generation model (cf. \S\ref{sec:framework:scaffolded:finetuned}) for all communities in Reddit. For that, we additionally prefix each scaffold with a line \texttt{subreddit: \{subreddit\}}. The model is trained for 60 epochs, using an Adafactor optimizer~\citep{shazeer2018adafactor}, a learning rate of 1e-3, and a dropout probability of $0.1$.

\subsection{Evaluation Methodology}
\label{sec:expeval:methodology}

We perform data generation on the community level: there are 250 subreddits as communities in our Reddit dataset, while the entire WikiPedia Talk Pages dataset represents a single community.
In each community, the discussion threads are randomly divided into a 50/50 test-train split. 
A sample of $N=50$ threads is drawn from the train portion as training examples. Based on this input, we attempt to generate $M=500$ synthetic discussion threads. 
This setting aims to measure the potential of our approaches to generate a significantly larger volume of synthetic data from a limited amount of real data.
The generated synthetic threads are then compared against the held-out test portion using the measures presented in Section~\ref{sec:evalmeasures}. 
Note that LLM-based generation does not always yield a valid thread, which is measured by \emph{success rate}.
Only valid threads are considered in the computation of the other measures.

\section{Results}
\label{sec:expeval}

We report results on the two collections separately: Reddit in Tables~\ref{tbl:res:reddit:struct} and \ref{tbl:res:reddit:content}, and Wikipedia Talk Pages in Tables~\ref{tbl:res:wikitalk:struct} and \ref{tbl:res:wikitalk:content}. 
In the case of Reddit, we report macro-averaged results across all communities (standard deviations are shows in parentheses).

Note that our measures vary in nature and should be interpreted accordingly.
Structural measures are collection-level statistics and are absolute values; they are to be compared against the same values computed on the test set.
Topic and content measures directly compare synthetic against real data; we also compute these on the training data (which the model had access to) to establish an upper limit on the performance a perfect method could achieve.
Finally, the realism measure is also an absolute value, which needs to be compared against real data for reference; for simplicity, we use the training set the model had access to for this.

Recall that our data generation framework is composed of three main processes (cf. Fig.~\ref{fig:pipeline}), and we have presented a number of alternatives for each of these.
For \emph{topic sampling}, we consider independent (IND) and conditional (COND) sampling algorithms.
For \emph{thread generation}, we have: 
\begin{itemize}

\item BASELINE: It can be zero-shot or few-shot; given that the zero-shot model had a very low success rate at producing valid threads ($<0.01$), we only report on the few-shot setting.
\item SCAFFOLD/FewS: Few-shot scaffold generation, followed by content generation that can be either zero-shot (CONT/ZeroS) or few-shot (CONT/FewS).
\item SCAFFOLD/FineT: Fine-tuned scaffold generation, followed by either zero-shot (CONT/ZeroS) or few-shot content generation (CONT/FewS).  As this model aims for cross-community generalization within a discussion platform, it is used exclusively on the Reddit dataset.
\end{itemize}
Note that, when it comes to structural measures, the specifics of the content-generation approach do not matter, as it does not have an impact on the resulting thread structure.

\paragraph{Validity}
Across both datasets, the success rate of the few-shot and fine-tuned scaffolding approaches greatly outperformed the baseline approach of generating the thread all-at-once. The reported confidence intervals do not overlap, indicating a statistically significant difference.
This confirms our basic premise that injecting LLM prompts with scaffolds is necessary for generating realistic synthetic UGC. Surprisingly, the threads produced from fine-tuned scaffolding were as successful as threads produced from few-shot scaffolding. This means that our fine-tuned model was able to generate valid threads at the same rate as a few-shot-prompted frozen model. The fact that the fine-tuned model can achieve this with only the topic as input shows that LLMs can learn valid thread structures, which is a promising insight for future work in generative language.

\paragraph{Structure}
For both datasets, the few-shot and fine-tuned scaffolding approaches tend to better capture the structure of real discussion threads---both in terms of graph structure (see Tables~\ref{tbl:res:reddit:struct} and \ref{tbl:res:wikitalk:struct}) as well as when comparing user-posting behavior (see Tables~\ref{tbl:res:reddit:user} and \ref{tbl:res:wikitalk:user} in Appendix~\ref{app:user_measures}). However, discrepancies observed across different metrics are not surprising,\footnote{Significant discrepancies are an artifact of strongly skewed power law distributions observed across structural measures computed on real discussion threads (e.g., on the test set, $\mathcal{W}$ takes values up to > 200M (on threads with extremely long discussion paths), while 90\% of test threads assume a value of < 1.2k).} given that our methodology does not explicitly model the underlying graph structure. In light of this, LLMs fare surprisingly well at reproducing specific aspects of graph structure, although there is room for improvement. This would be an interesting direction to explore in the future.

\paragraph{Topics}
For each generation approach, we measure the topical similarity to the ground-truth test set by two distributional measures, Jensen-Shannon (J-S), and Weighted Jaccard (W-J) (cf. Section~\ref{sec:evalmeasures:topic}). First, note that even when we compare the train set to the test set, these similarities are only at 0.76 and 0.72 (respectively), representing an upper-bound of similarity that any approaches can achieve. Across both datasets, we find that topic similarity slightly approaches the upper-bound as we move from the baseline approach, to few-shot scaffolding, to the fine-tuned scaffolding approach. 
We find all differences between the baseline and non-baseline generators to be statistically significant (at $p<0.05$), however differences between the non-baseline generators are not statistically significant. There are opportunities for future work fine-tuning LLMs on real discussion towards faithfully replicating the discussed topics.

\paragraph{Content}
We further compare embeddings corresponding to content generated from synthetic and real data using MAUVE via different settings. As observed from tables~\ref{tbl:res:reddit:content} and~\ref{tbl:res:wikitalk:content}, the upper limits on the performance a perfect method can obtain are always achieved on training data. We observe that the next best performances across both Reddit and Wikipedia Talk Pages datasets are achieved via Fine-tuned scaffolding (SCAFFOLD/FineT) using zero-shot (CONT/ZeroS) or few-shot content generation(CONT/FewS). Additionally, COND topic sampling algorithm often performed better than IND topic sampling algorithm.    

\paragraph{Realism}
In general, our methods scored highly on our novel LLM-based discussion realism metric. Interestingly, the fine-tuned scaffold generation approach registered consistently lower by about 10\%. We hypothesize that this is due to gaps in the ability of our fine-tuned model to generate useful summaries for zero-shot/few-shot content generation. This is a promising area for future work.

\paragraph{Summary of findings}
Overall, we find our approaches to be effective at generating synthetic data that is similar to real data (RQ1). In terms of more fine grained methodological aspects, the more sophisticated topic sampling performs marginally better with minimal differences (RQ2). Scaffolding-based generation, however, shows very clear benefits in generating valid thread structure (RQ3). Our results show good generalization across different types of online communities, although the benefits of fine-tuning a single model for all communities within a platform only marginally outperforms simply picking few-shot examples from each community (RQ4).

\paragraph{Recitation analysis}
The recitation of training data (either from the pre-training corpus or from the few-shot examples) is a valid concern (noting that is a more generic issue with generative AI that is not unique to our problem). We performed a check for close matches between synthetic and real threads and found those to be extremely rare; we refer to Appendix~\ref{app:recitation} for the details.

\paragraph{Cost}
Using a PaLM 2 model served on Google Cloud\footnote{\url{https://cloud.google.com/vertex-ai/generative-ai/pricing}} the cost of topic extraction for the 250 subreddits in our experiments is \$7.5, and generating 500 threads per subreddit using the few-shot scaffolding approach costs about \$975. This means that a reasonably large scale synthetic collection, comprising 125k threads and around 1.5M posts in total, can be created under \$1000.
\section{Conclusion}
\label{sec:conclusion}

In this paper, we have addressed the challenge of generating realistic synthetic discussion threads using a multi-step pipeline by leveraging LLMs. 
We have also introduced a suite of evaluation measures designed to comparing synthetic and real threads in terms of structure, topics, content, and realism. Our results on two online discussion platforms with markedly different characteristics demonstrate that our approaches are capable of generating synthetic data meaningfully from a small sample of real data.

While these results are promising, this work represents an initial attempt at synthetic discussion thread generation, and there remain numerous avenues for future research. 
This includes potentially identifying other ways to explicitly encode thread structure, which proved particularly valuable in our results, on top of determining optimal approaches for designing prompts and both the number and type of examples used.

Another line of future investigation concerns evaluation measures. Our suite of metrics provides a starting point, but is by no means exhaustive.
We have considered manual annotation of the generated data, but it is challenging for two reasons. First, it proved difficult to set up a meaningful annotation task and to identify what attributes of discussion threads should be annotated. Second, recent studies indicate that humans have difficulty reliably distinguishing between real and AI-generated text~\citep{Casal:2023:RMAL,Jakesch:2023:PNAS}. This casts doubt on the potential effectiveness of manual annotation for our case.

In general, there is also an inherent trade-off between privacy and utility when machine learning models are trained on sensitive data. Synthetic data generators in particular present a risk of potentially encoding unexpected private information as they learn the original data distribution. Solutions to protect the original data include enforcing privacy at inference time and differential privacy training~\cite{abadi2016deep,kurakin2023harnessing}. 
Application of these specifically to synthetic UGC is also a valuable avenue of future research.

Finally, a comprehensive study of the utility of this synthetic data across various applications, including its potential use for data augmentation, is planned to be explored in future work, along with a comprehensive discussion of limitations, root causes, and potential solutions.

\bibliographystyle{ACM-Reference-Format}
\bibliography{references}

\appendix

\section{Data}
\label{app:dataset}

We provide details regarding data preprocessing and sampling.

\subsection{Reddit}
\label{app:dataset:reddit}

We first build all threads from all subreddits, ignoring any thread marked 
``not-safe-for-work'' by the Reddit platform (a field which comes with the Pushshift data), and removing any sub-thread with a root post marked deleted or removed (if the top-level post is removed, the whole thread is ignored). We note that a uniform sample of threads would have low-diversity, since the majority of threads on Reddit have few or no replies, and a large number simply reshare a link or an image. To arrive at a useful set of threads, we rely on a form of stratified sampling: First, we restrict the set of subreddits to those with at least 100 threads, 100 comments, and 10 unique active users during April 2019. We then compute four metrics on each subreddit: average thread length (by number of posts), average thread depth (longest reply chain), average number of characters in the thread (across all posts), and average initial post number of characters. We compute four separate four-way partitions of the subreddits, each partition grouping the subreddits by one of the four metrics. We then sample eight subreddits uniformly-at-random from each of the 16 partition groups. This produced 250 unique subreddits with high diversity of discussion lengths by various measurements.

\subsection{Wikipedia Talk Pages}
\label{app:dataset:wikitalk}

Wikipedia Talk Pages discussions contain post/reply modification, deletion, and restoration actions, which are explicitly recorded in the user-facing data. In our experiments, we take the ``final form'' of each conversation to extract a discussion tree, which ignores these edit-type actions. In many cases, the final form of the conversation is not a valid fully-connected thread, due to the original parent IDs of replies being modified or removed after edit actions. We remove all such irreconcilable discussions from the dataset, arriving at 1.3 million valid threads with greater than one post and maximum thread size 217. The vast majority (1.1M) of these threads have only 2 posts: an initial post, and a reply. We therefore stratify-sample threads by their size to obtain a balanced dataset. Let $\mathcal{T}_m$ be the set of WikiConv threads of size $m$. Define $n_m = 300 + 1(|\mathcal{T}_m| > 1000) * 200$. We sample $\min(|\mathcal{T}_m|, n_m)$ threads from $\mathcal{T}_m$ uniformly-at-random, and use the (disjoint) union of these samples over $m$ to construct our dataset.

\subsection{User-specific Measures}
\label{app:user_measures}

In addition to evaluating the structure of the synthetically generated threads in comparison to real threads, we also investigate the corresponding user-posting behavior. To this end, individual users are identified based on the user IDs associated with individual posts. Since our framework models within-thread (rather than across-thread) user behavior, user IDs corresponding to synthetically generated threads are only unique with respect to individual threads. To make the results comparable with the real data, we also preprocess user IDs found in the real threads accordingly.

The results for the Reddit and the Wikipedia Talk Pages datasets can be found in Tables~\ref{tbl:res:reddit:user} and~\ref{tbl:res:wikitalk:user}, respectively, and are in line with the observations described in the main text.

\begin{table*}[t]
    \caption{Thread generation results on the Reddit dataset in terms of metrics evaluating within-thread per-user behavior. Measures are macro-averages over subreddits, with standard deviations shown in parentheses. The first row reports these statistics on the test set for reference comparison.}
    \label{tbl:res:reddit:user}
    \resizebox{0.7\textwidth}{!}{%
    \begin{tabular}{l|l|cccc}
    \toprule
    \textbf{Topic} & \textbf{Thread} & \multicolumn{4}{c}{\textbf{User-specific measures}} \\
    \textbf{sampl.} & \textbf{generation} & \textbf{\#posts} & \textbf{Mean post depth} & \textbf{Mean \#direct replies} & \textbf{Mean \#(in)direct replies} \\
    \midrule
        n/a & TEST SET & 1.59 ($\pm 0.64$) & 1.61 ($\pm 0.73$) & 0.74 ($\pm 0.21$) & 1.48 ($\pm 0.80$) \\
    \midrule
        \multirow{3}{*}{IND}
            & BASELINE & 1.15 ($\pm 0.19$) & 1.46 ($\pm 0.38$) & 0.84 ($\pm 0.05$) & 1.40 ($\pm 0.39$) \\
            & SCAFFOLD/FewS & 1.11 ($\pm 0.20$) & 1.66 ($\pm 0.29$) & 0.85 ($\pm 0.05$) & 1.64 ($\pm 0.29$) \\
            & SCAFFOLD/FineT & 1.43 ($\pm 0.45$) & 1.48 ($\pm 0.54$) & 0.74 ($\pm 0.10$) & 1.36 ($\pm 0.53$) \\
    \midrule
        \multirow{3}{*}{COND}
            & BASELINE & 1.15 ($\pm 0.19$) & 1.46 ($\pm 0.37$) & 0.84 ($\pm 0.05$) & 1.40 ($\pm 0.39$) \\
            & SCAFFOLD/FewS & 1.11 ($\pm 0.20$) & 1.67 ($\pm 0.29$) & 0.86 ($\pm 0.05$) & 1.65 ($\pm 0.30$) \\
            & SCAFFOLD/FineT & 1.44 ($\pm 0.49$) & 1.50 ($\pm 0.57$) & 0.75 ($\pm 0.10$) & 1.38 ($\pm 0.56$) \\
    \bottomrule
    \end{tabular}
    }
\end{table*}

\begin{table*}[t]
    \caption{Thread generation results on the Wikipedia Talk Pages dataset in terms of metrics evaluating within-thread per-user behavior. The first row reports these statistics on the test set for reference comparison.}
    \label{tbl:res:wikitalk:user}
    \resizebox{0.7\textwidth}{!}{%
    \begin{tabular}{l|l|cccc}
    \toprule
    \textbf{Topic} & \textbf{Thread} & \multicolumn{4}{c}{\textbf{User-specific measures}} \\
    \textbf{sampl.} & \textbf{generation} & \textbf{\#posts} & \textbf{Mean post depth} & \textbf{Mean \#direct replies} & \textbf{Mean \#(in)direct replies} \\    
    \midrule
        n/a & TEST SET & 10.60 & 2.49 & 0.91 & 2.73 \\
    \midrule
        \multirow{2}{*}{IND}
            & BASELINE & 1.75 & 1.83 & 0.73 & 1.45 \\
            & SCAFFOLD/FewS & 1.53 & 2.73 & 0.87 & 2.51 \\
    \midrule
        \multirow{2}{*}{COND}
            & BASELINE & 1.76 & 1.97 & 0.72 & 1.56 \\
            & SCAFFOLD/FewS & 1.51 & 2.42 & 0.85 & 2.19 \\
    \bottomrule
    \end{tabular}
    }
\end{table*}

\section{Realism Measure}
\label{app:realism}
The prompt design for our realism evaluation method relies on few-shot examples and chain-of-thought injection, as follows:
\begin{footnotesize}
\begin{verbatim}
You are an expert analyst of online discussions on forum sites like Reddit.
On forum sites, discussions are started by posts, and posts can receive
reply posts. Reply posts can also receive further reply posts. A 
"discussion path" is a sequence of replies starting with the initial post.

Sometimes Reddit reply posts are falsified or fake. When this happens, 
the discussion could become unrealistic or not coherent. Your job is to 
examine a given discussion path and determine whether it seems realistic 
and coherent.

Here is an example:

TITLE: {title of post}
POST[user-0]: {content of initial post}
REPLY[user-1]: {content of first reply}
REPLY[user-0]: {content of return reply}
...

EXPLANATION: {discussion_analysis}
ANSWER: The answer is {yes,no}.

{target_example}
\end{verbatim}
\end{footnotesize}
The discussion analysis in the explanation field is the chain-of-thought exposition of the discussion. We provide the LLM examples of both coherent and incoherent discussions, populating the explanation field with formal descriptions of the discussion and its (in)coherence. The target example is written in the same format as the few-shot examples.

\begin{figure}[t]
    \centering
    \includegraphics[width=0.5\linewidth]{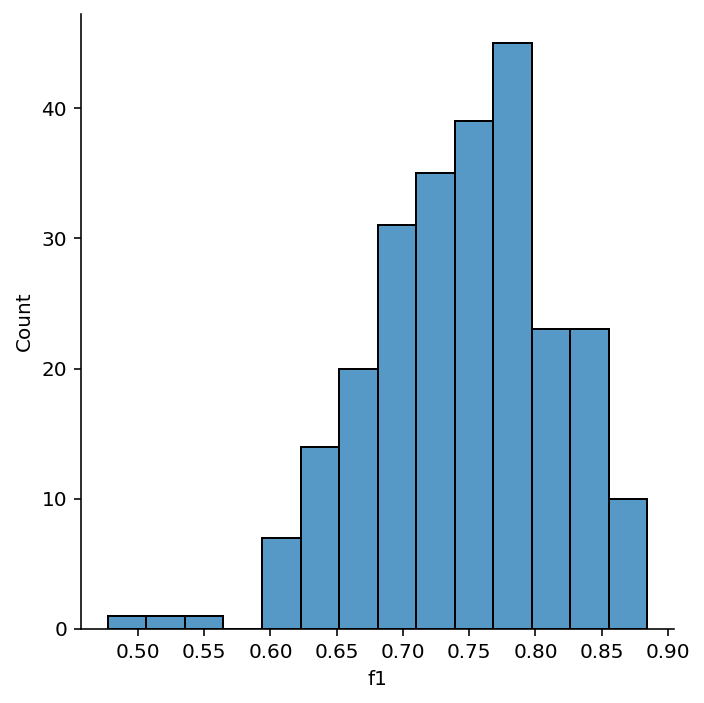}
    \caption{F1 scores across subreddits of the discussion-path realism prompting scheme. The score is computed on a balanced sample of real discussion-paths and ``corrupted'' discussion paths with some randomly-chosen posts replaced with randomly-chosen posts from the rest of the subreddit.}
    \label{fig:f1-realism}
\end{figure}

To provide a meta-evaluation of our prompt design for the realism evaluation, we treat the prompted LLM as a binary classifier, and generate negative examples to evaluate its classification accuracy. First, we sample $N=100$ paths from each subreddit, and then we sample $M=1$ path from each thread as a positive example. For each path, we create a corresponding negative example path by choosing at least one post from the path to swap with another post randomly sampled from our corpus. We control for (1) topical and (2) post-length effects (retaining a focus on discussion-coherence effects) by sampling replacement posts (1) from the same subreddit and (2) from the same depth as the original post. This creates a balanced dataset of 
~25,000 positive examples and ~25,000 negative examples (some subreddits do not have 100 posts during the time interval of our dataset). We apply our prompt scheme to the augmented dataset and compute the F1 scores for each subreddit, shown in Fig.~\ref{fig:f1-realism}.

Fig.~\ref{fig:f1-realism} shows that our realism evaluation is good, but not perfect for every subreddit, and near-random for a few. This is likely due to some subreddits being less obviously ``conversational'' than others: users may not directly reply to the parent comment, instead uttering short phrases, links, slang, or offhand irrelevant comments.

\section{LLM Prompts}
\label{app:prompts}

Below we show specific instantiations of our general prompt structure (cf. Section~\ref{sec:expsetup:llms}) that we use in various components of our pipeline.
Input variables are in curly brackets.

\subsection{Topic Extraction}
\label{app:prompts:topics}

\begin{footnotesize}
\begin{verbatim}
Extract topics from social media threads. 
Here are some examples:

Here is a thread:
{example thread 1}
Here are the topics for the thread:
{topics for thread 1}

Here is a thread:
{input thread}
Here are the topics for the thread:
\end{verbatim}
\end{footnotesize}

\subsection{Summarization}
\label{app:prompts:summaries}

\begin{footnotesize}
\begin{verbatim}
You are summarizing social media posts. The summary should be written 
in third person, starting with "The user".

<repeat for the number of examples>
Here is an example:
POST:
{example post}
SUMMARY:
{example summary}
</repeat>

Now, summarize the following post:
POST:
{input post}
SUMMARY:
\end{verbatim}
\end{footnotesize}

\subsection{Baseline Thread Generation}
\label{app:prompts:baseline}

Below is the prompt used for Reddit. For Wikipedia Talk Pages the prompt is the same except the first line (``...single thread of conversations that happens on Wikipedia Talk Pages.") \\

\begin{footnotesize}
\begin{verbatim}
Given a list of topics, create a single thread of Reddit-like 
conversations.

The conversation should be represented as follows:
The first line starts with "title:" followed by the title of the 
discussion thread. This is followed by n lines, each corresponding 
to a post in the thread. There is always at least an initial post, 
which may be followed by comments from users. Users can comment on 
the original post or any of the previous comments. Users can make 
multiple comments, but each comment should specify which previous 
comment it is a reply to.

Each post should include the following four fields, separated by 
hashtags (#):

1. The ID of the post. The value is either "post", if it is the 
   initial post, or "comment-X" (where X is 1, 2, 3, etc.)
2. The ID of the user who wrote the post. The value is "user-Y" 
   (where Y is 1, 2, 3, etc.) The initial post is always written 
   by "user-1".
3. The ID of the parent post. The value is "NA" if it is the 
   initial post, "post" if it is a comment to the initial post, or
   "comment-X" is it is a reply to one of the previous comments.
4. The content of the post.

<repeat for the number of examples>
Here is an example:
TOPICS:
{example topics}
THREAD:
{example thread}
</repeat>

Now, create a thread for the following:
TOPICS:
{input topics}
THREAD:
\end{verbatim}
\end{footnotesize}

\subsection{Few-shot Scaffold Generation}
\label{app:prompts:scaffolds}

Below is the prompt used for Reddit. For Wikipedia Talk Pages the prompt differs in the first line as explained above.\\

\begin{footnotesize}
\begin{verbatim}
Given a list of topics, create a single thread of a Reddit-like
conversation.

The conversation should be represented as follows:
The first line starts with "title:" followed by the title of the 
discussion thread. This is followed by n lines, each corresponding 
to a post in the thread. There is always at least an initial post, 
which may be followed by comments from users. Users can comment on 
the original post or any of the previous comments. Users can make 
multiple comments, but each comment should specify which previous 
comment it is a reply to.

Each post should include four fields separated by hashtags (#):

1. The ID of the post, which is "post" for the initial post and 
   "comment-X" for the following comments, where X is 1, 2, 3, etc.
2. The ID of the user who wrote the post, following the format 
   "user-Y", where Y is 1, 2, 3, etc. The initial post is always 
   written by "user-1".
3. Whether the comment is a reply to the original post or to one 
   of the previous comments. If it's a reply to the original post,
   this field should contain the text "post". If it's a reply to
   a previous comment, it should contain the keyword "comment" 
   followed by the number of the comment it's replying to. In 
   case of the initial post, this field always has the value "NA".
4. The content of the post. It should be a somewhat coherent 
   sentence expressing the intent of the user, written in third 
   person and starting with "The user".

<repeat for the number of examples>
Here is an example:
TOPICS:
{example topics}
THREAD:
{example thread}
</repeat>

Now, create a thread for the following:
TOPICS:
{input topics}
THREAD:
\end{verbatim}
\end{footnotesize}

\subsection{Content Generation}
\label{app:prompts:content}

Prompt for generating the initial post of the thread: \\

\begin{footnotesize}
\begin{verbatim}
You are writing the first post of a thread on a discussion forum.

<repeat for the number of examples>
Here is an example of how to write the post based on the summary.
The summary of the post is:
{example summary}
The post text is:
{example post from summary}
</repeat>

The thread will discuss the following topics:
{input topics}
The summary of the post is:
{input summary}
The post text is:
\end{verbatim}
\end{footnotesize}

\vspace{\baselineskip}
\noindent
Prompt for generating the content of replies in the thread: \\

\begin{footnotesize}
\begin{verbatim}
You are writing the next post of a thread on a discussion forum.

<repeat for the number of examples>
Here is an example of how to write the post based on the summary.
The summary of the post is:
{example summary}
The post text is:
{example post from summary}
</repeat>

The thread discusses the following topics:
{input topics}
The thread so far is:
<repeat for each post in the thread so far>
{user}: {post content}
</repeat>
The summary of the post is:
{input summary}
The post text is:
\end{verbatim}
\end{footnotesize}

\section{Recitation Analysis}
\label{app:recitation}

We performed the following analysis to check whether text in the generated data appears in the training data. The main goal is to check that, for a particular subreddit $s$, there are no non-trivial duplicated discussions between the real threads $\mathcal{T}_s$ and a given batch of synthetic threads $\widehat{\mathcal{T}}_s$. If we simply did pairwise comparisons of all comments, we would likely find exact matches for trivial comments such as ``\emph{Thank you!}.'' To mitigate this issue, we produced T5 encodings of concatenations of each comment with its parent. We then computed all pairwise cosine similarities between these comment representations from $\mathcal{T}_s$ and $\widehat{\mathcal{T}}_s$. We found only a few exact matches from trivial first-level reply comments on threads with no top-post text, with comment text such as ``\emph{Are you selling this?}'' and ``\emph{bruh}.'' Other high-similarity pairs came from $\mathcal{T}_s$ comments that had real links (e.g., from YouTube) and $\widehat{\mathcal{T}}_s$ comments that had hallucinated links from the same domain name. Overall, the cosine similarity distribution .999 quantile was $\approx$0.93, which tells us that close matches between our synthetic threads and the real threads are extremely rare.

\end{document}